\documentclass[conference]{IEEEtran}
\IEEEoverridecommandlockouts
\usepackage{cite}
\usepackage{hyperref}
\usepackage{amsmath,amssymb,amsfonts}
\usepackage{algorithmic}
\usepackage{graphicx}
\usepackage{textcomp}
\usepackage{xcolor}
\usepackage{url}
\def\BibTeX{{\rm B\kern-.05em{\sc i\kern-.025em b}\kern-.08em
    T\kern-.1667em\lower.7ex\hbox{E}\kern-.125emX}}
\begin{document}

\title{EvoMan: Game-playing Competition}

\author{\IEEEauthorblockN{Fabricio Olivetti de Franca, Denis Fantinato}
\IEEEauthorblockA{\textit{Heuristics, Analysis and Learning Laboratory (HAL)} \\
\textit{Center of Mathematics, Computing and Cognition (CMCC)}\\
\textit{Federal University of ABC}\\
Santo Andre, Brazil \\
\{folivetti, denis.fantinato\}@ufabc.edu.br}
\and
\IEEEauthorblockN{Karine Miras, A.E. Eiben}
\IEEEauthorblockA{\textit{Department of Computer Science} \\
\textit{Vrije Universiteit}\\
Amsterdam, Netherlands \\
karine.smiras@gmail.com}
\and
\IEEEauthorblockN{Patricia A. Vargas}
\IEEEauthorblockA{\textit{Edinburgh Centre for Robotics} \\
\textit{Heriot-Watt University}\\
Edinburgh, UK \\
P.A.Vargas@hw.ac.uk}
\and

}

\maketitle

\begin{abstract}
This paper describes a competition proposal for evolving Intelligent Agents for the game-playing framework called EvoMan. The framework is based on the boss fights of the game called Mega Man II developed by Capcom. For this particular competition, the main goal is to beat all of the eight bosses using a generalist strategy. In other words, the competitors should train the agent to beat a set of the bosses and then the agent will be evaluated by its performance against all eight bosses. At the end of this paper, the competitors are provided with baseline results so that they can have an intuition on how good their results are.
\end{abstract}


\section{Introduction}
EvoMan~\cite{karinemiras01} is a framework for testing competitive game-playing agents in a number of distinct challenges such as:

\begin{itemize}
\item Learning how to win a match against a single enemy
\item  Generalizing the agent to win the matches against the entire set of enemies
\item  Coevolving both the agent and the enemies to create intelligent enemies with increasing difficulties.
\end{itemize}

This framework is inspired on the boss levels of the game Mega Man II~\cite{capcom} created by Capcom in which the player controls a robot equipped with a simple arm cannon and must beat $8$ Robot Masters equipped with different weapons.

In the game, every time Mega Man defeats a Robot Master, it acquires its weapon making it easier to defeat the remainders bosses. This game is considered to have a high degree of difficulty among skilled players. As a personal challenge, some skilled players try to beat all Robot Masters using only the default arm cannon.

With this competition we propose the following question: can a fully automated intelligent agent defeat each one of the Robot Masters using only the default arm cannon? In more details, we challenge the competitors to evolve an intelligent agent by allowing it to train against four ($4$) of the Robot Masters but with the ultimate goal of defeating all eight ($8$) of them.

\section{The Challenge}

In this challenge, the contestants should train their agent on a set of four enemies (defined by the contestant) and evaluate how general is their learned strategy when fighting against the whole set of enemies.

Since each enemy behavior greatly differs from each other, the player should learn how to identify and react to general patterns like avoiding being shot or shoot at the direction of the enemy. Learning a general strategy capable of winning over the entire enemies set can be very challenging~\cite{karinemiras01,karinemiras02}. 

\begin{figure}
    \centering
    \includegraphics[width=0.4\textwidth]{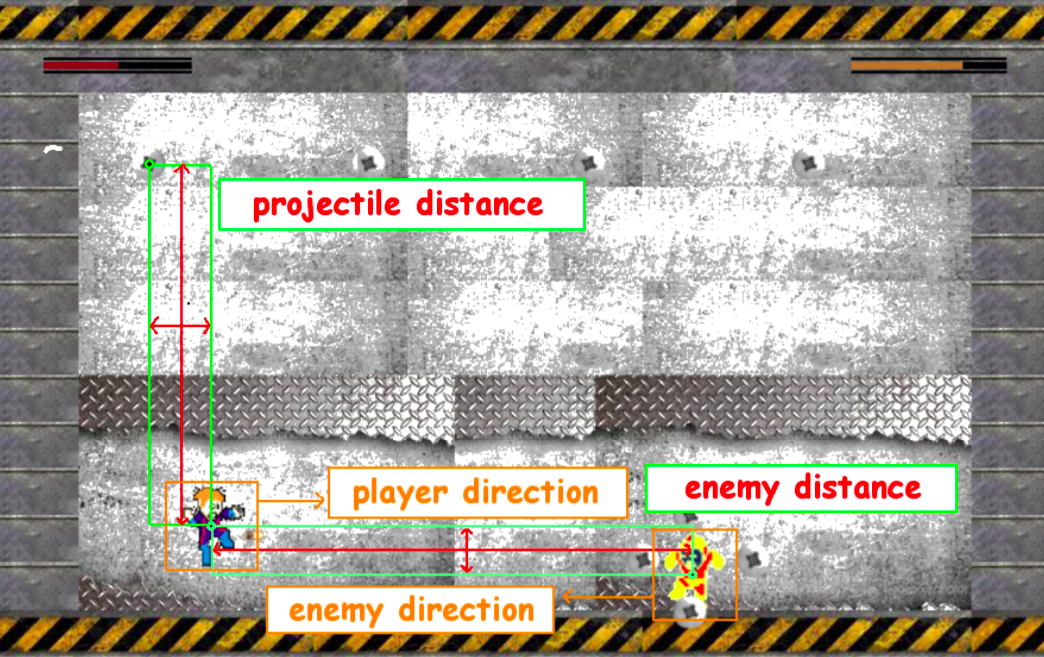}
    \caption{Sensors available for the competition.}
    \label{fig:sensors}
\end{figure}

The agent will have a total of $20$ sensors, with $16$ of them corresponding for horizontal and vertical distance to $8$ different bullets (maximum allowed), $2$ to the horizontal and vertical distance to the enemy, and $2$ describing the direction the player and the enemy is facing. The sensors are illustrated in Fig.~\ref{fig:sensors}.

The framework is freely available\footnote{\url{https://github.com/karinemiras/evoman_framework}} and it is currently compatible with Python 3.6 and 3.7 (Python 3.8 is not compatible at the moment). There is also an extensive documentation available\footnote{\url{https://github.com/karinemiras/evoman_framework/blob/master/evoman1.0-doc.pdf}}.

\section{Evaluation Criteria}

Both the agent and the enemies start the game with $100$ energy points. Every time the player or the enemy gets hit, they lose one point. Whoever reaches $0$ points loses the match.
The final performance of the agent after the end of a match is calculated by the energy gain, as a maximization problem, calculated by the difference between the player and the enemy energy:

\begin{equation*}
Gain = 100.01 + ep - ee,    
\end{equation*}

\noindent where $ee$ and $ep$ are the final amount of energy of the enemy and the player, respectively. The value of $100.01$ is added so that the harmonic mean always produces a valid result.

The main goal of this competition is that a given agent perform equally good for every boss. So, each contestant agent will be tested against all of the enemies, and they will be ranked by the harmonic mean of the performance over the different bosses.

\section{Participating in the Competition}

The initial code, manual and every other needed resources are available at a Github repository\footnote{\url{https://github.com/karinemiras/evoman_framework}}. The competitors should pay attention to the following directions:

\begin{itemize}
    \item Follow the installation instructions in the file evoman1.0-doc.pdf.
    \item Run the demo script controller\_specialist\_demo.py to test if the framework is working.
    \item Play the game using your own keyboard to understand the difficulties of the problem. Use the script \emph{human\_demo.py} for that.
    \item The agent should be trained using the Individual Evolution and Multi-objective modes with the goal of beating each one of the four adversaries chosen for training.
\end{itemize}

\section{Current Results}

In~\cite{karinemiras03} different learning strategies were tested for the individual evolution mode. In this mode the algorithm creates one agent for each boss, thus generating specialist agents. The algorithms used for this test was variants of neuroevolution~\cite{neuro} strategies with $1$-layer perceptron and $2$-layers perceptron with $10$ and $50$ neurons for the hidden layer. The weights of the Neural Network was adjusted by means of a Genetic Algorithm~\cite{ga} (GAP, GA10, GA50) and LinkedOpt algorihtm~\cite{linkedopt} (LOP, LO10, LO50). The other strategy was the evolution of a Neural Network topology with their weights by means of the NEAT algorithm~\cite{neat}.

Table~\ref{tab:results} show the obtained results so far for the individual mode. These results serve as an upper bound for the proposed competition.

From this table we can see that NEAT provided the best overall results followed by the two-layer neural networks with their weights adjusted through a Genetic Algorithm.

Notice that since these results were obtained with Individual Mode they will most likely serve as an upper bound of the results obtained in this  competition.

\begin{table}[t!]
    \caption{Gains obtained by each tested algorithm reported in~\cite{karine03}. The \emph{mean} row is the harmonic mean of the results.}
    \centering
    \begin{tabular}{c|ccccccc}
    \hline
        Boss &  NEAT  &  GAP  & GA10  & GA50 & LOP   & LO10  & LO50 \\
        \hline
         1    & 190.01 & 190.01 & 190.01 & 190.01 & 0.01   & 196.01 & 80.01  \\ 
2    & 194.01 & 190.01 & 182.01 & 178.01 & 190.01 & 182.01 & 188.01 \\ 
3    & 180.01 & 158.01 & 158.01 & 136.01 & 124.01 & 70.51  & 116.01 \\ 
4    & 194.01 & 93.51  & 118.01 & 169.01 & 73.01  & 36.51  & 119.01 \\ 
5    & 194.01 & 180.01 & 188.01 & 179.01 & 178.01 & 181.01 & 188.01 \\ 
6    & 173.01 & 79.01  & 77.51  & 103.01 & 139.01 & 128.01 & 20.01  \\ 
7    & 177.01 & 170.01 & 156.01 & 118.01 & 186.01 & 169.01 & 190.01 \\ 
8    & 186.01 & 177.01 & 183.01 & 178.01 & 0.01   & 182.01 & 183.01 \\ 
Mean & 185.67 & 139.64 & 143.74 & 149.43 & 0.04   & 104.01 & 79.32  \\ 
         \hline
    \end{tabular}
    \label{tab:results}
\end{table}

\section{Conclusion}

This paper introduces the EvoMan competition as a new general game playing challenge based on the Mega Man II game by CAPCOM. The main goal of the competition is to evolve an intelligent agent capable of defeating each one of the eight available bosses while being able to train against a smaller subset of those enemies.

The winner of this competition will be the one agent that performs equally well on each one of the eight bosses, hopefully defeating them all.

In order to help the competitors, we provided a table with the upper bounds of the gains for each boss, obtained with the help of specialist agents.


\end{document}